\documentclass[runningheads]{llncs}
\usepackage{times}
\usepackage{epsfig}
\usepackage{graphicx}
\usepackage{amsmath}
\usepackage{amssymb}
\usepackage{bm}
\usepackage{nicefrac}
\usepackage{booktabs}
\usepackage{multirow}
\usepackage{tabulary}
\usepackage{microtype}
\usepackage{caption}
\usepackage{subcaption}
\usepackage{xspace}
\usepackage[inline]{enumitem}
\usepackage{csquotes}
\usepackage{wasysym}

\usepackage[pagebackref=true,breaklinks=true,colorlinks,bookmarks=false]{hyperref}

\usepackage{cleveref}
\usepackage{xcolor}

 \newcommand\blfootnote[1]{%
   \begingroup
   \renewcommand\thefootnote{}\footnote{#1}%
   \addtocounter{footnote}{-4}%
   \endgroup
 }

\newcommand{\myparagraph}[1]{\vspace{4pt}\noindent{\bf #1}}

\newcommand{\eg}{e.g.\ }

\newcommand{\rev}[1]{\textcolor{black}{#1}}

\newcommand{\clevrx}{CLEVR-X\xspace}
\newcommand{\vqax}{VQA-X\xspace}
\newcommand{\esnlive}{e-SNLI-VE\xspace}

\begin{document}
\title{CLEVR-X: A Visual Reasoning Dataset for Natural Language Explanations}
%
%
\author{Leonard Salewski\inst{1}\orcidID{0000-0001-8531-3011} \and
A. Sophia Koepke\inst{1}\orcidID{0000-0002-5807-0576} \and
Hendrik P.\ A.\ Lensch\inst{1}\orcidID{0000-0003-3616-8668} \and
\mbox{Zeynep Akata}\inst{1,2,3}\orcidID{0000-0002-1432-7747}}
\authorrunning{L. Salewski et al.}
%
\institute{University of T{\"u}bingen, T{\"u}bingen, Germany \and
MPI for Informatics, Saarbrücken, Germany \and
MPI for Intelligent Systems, T{\"u}bingen, Germany\\
\email{\{leonard.salewski, a-sophia.koepke, hendrik.lensch, zeynep.akata\}@uni-tuebingen.de}}
\maketitle              
\begin{abstract}
Providing explanations in the context of Visual Question Answering (VQA) presents a fundamental problem in machine learning. 
To obtain detailed insights into the process of generating natural language explanations for VQA, we introduce the large-scale CLEVR-X dataset that extends the CLEVR dataset with natural language explanations. For each image-question pair in the CLEVR dataset, CLEVR-X contains multiple structured textual explanations which are derived from the original scene graphs. By construction, the CLEVR-X explanations are correct and describe the reasoning and visual information that is necessary to answer a given question. We conducted a user study to confirm that the ground-truth explanations in our proposed dataset are indeed complete and relevant.
We present baseline results for generating natural language explanations in the context of VQA using two state-of-the-art frameworks on the CLEVR-X dataset. Furthermore, we provide a detailed analysis of the explanation generation quality for different question and answer types.
Additionally, we study the influence of using different numbers of ground-truth explanations on the convergence of natural language generation (NLG) metrics. The CLEVR-X dataset is publicly available at \url{https://explainableml.github.io/CLEVR-X/}.
\end{abstract}
 \section{Introduction}

Explanations for automatic decisions form a crucial step towards increasing transparency and human trust in deep learning systems. In this work, we focus on natural language explanations in the context of vision-language tasks.
\blfootnote{A version of the contribution has been accepted for publication, after peer review but is not the Version of Record and does not reflect post-acceptance improvements, or any corrections. The Version of Record will be available online at: \url{https://doi.org/10.1007/978-3-031-04083-2_5}.}

In particular, we consider the
vision-language task of Visual Question Answering (VQA) which consists of answering a question about an image. This requires multiple skills, such as visual perception, text understanding, and cross-modal reasoning in the visual and language domains. A natural language explanation for a given answer allows a better understanding of the reasoning process for answering the question and adds transparency. 
However, it is challenging to formulate what comprises a good textual explanation in the context of VQA involving natural images.

\begin{figure}[t]
    \centering
    \begin{subfigure}[t]{0.31\textwidth}
        \centering
        \scriptsize
        \caption*{%
        {\bf VQA-X} \\ \textbf{Question}: Does this scene look like it could be from the early 1950s?}
        \includegraphics[width=0.75\linewidth,trim={0 3mm 0 1.5mm},clip]{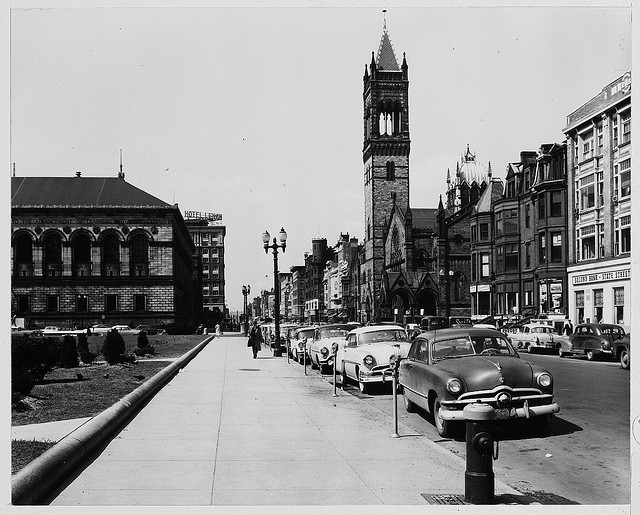}
        \caption*{\textbf{Answer $\mid$ Explanation:}\\
        Yes $\mid$ The photo is in black and white and the cars are all classic designs from the 1950s
}%
       \label{fig:vqax-example}
    \end{subfigure}
    \hfill
    \begin{subfigure}[t]{0.31\textwidth}
       \centering
        \caption*{{\bf e-SNLI-VE} \\ \textbf{Hypothesis}: A woman is holding a child.\\}
        \includegraphics[width=0.75\linewidth,trim={0 6mm 0 6mm},clip]{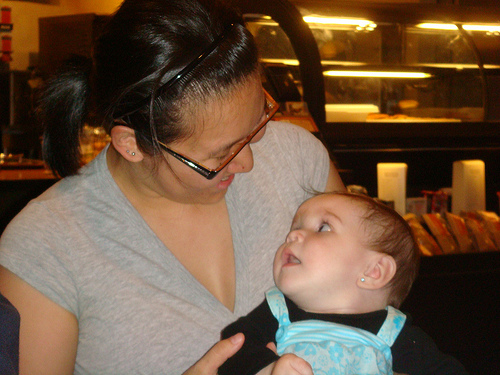}
        \caption*{\textbf{Answer $\mid$ Explanation:}\\
        Entailment $\mid$ If a woman holds a child she is holding a child.}%
        \label{fig:esnlive-examples}
    \end{subfigure}
     \hfill
    \begin{subfigure}[t]{0.31\textwidth}
        \centering
        \caption*{{\bf CLEVR-X} \\\textbf{Question}: There is a purple metallic ball; what number of cyan objects are right of it?}
        \includegraphics[width=0.75\linewidth]{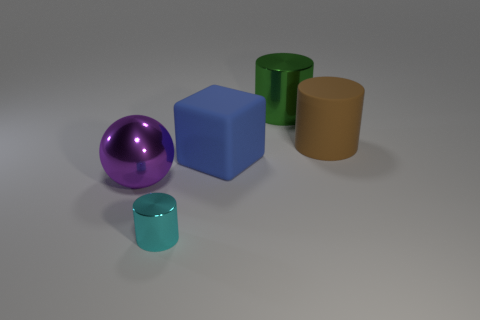}
        \caption*{\textbf{Answer $\mid$ Explanation:}\\
        1 $\mid$ There is a cyan cylinder which is on the right side of the purple metallic ball.}%
       \label{fig:clevrx-examples}
    \end{subfigure}
    \caption{Comparing examples from the \vqax{} (left), \esnlive{} (middle), and \clevrx{} (right) datasets. The explanation in \vqax{} requires prior knowledge (about cars from the 1950s), \esnlive{} argues with a tautology, and our \clevrx{} only uses abstract visual reasoning.}%
    \label{fig:teaser-dataset-comparison}
\end{figure}

Explanation datasets commonly used in the context of VQA, such as the VQA-X dataset~\cite{hukpark2018MultimodalExplanationsJustifying} or the e-SNLI-VE dataset~\cite{Do2020eSNLIVE20CV,kayser2021vil} for visual entailment,
contain explanations of widely varying quality since 
they are generated by humans. The ground-truth explanations in VQA-X and e-SNLI-VE can range from statements that merely describe an image to explaining the reasoning about the question and image involving prior information, such as common knowledge. One example for a ground-truth explanation in VQA-X that requires prior knowledge about car designs from the 1950s can be seen in Fig.~\ref{fig:teaser-dataset-comparison}. The e-SNLI-VE dataset contains numerous explanation samples which consist of repeated statements (``x because x'').
Since existing explanation datasets for vision-language tasks contain immensely varied explanations, it is challenging to perform a structured analysis of strengths and weaknesses of existing explanation generation methods.

In order to fill this gap, we propose the novel, diagnostic CLEVR-X dataset for visual reasoning with natural language explanations. It extends the synthetic CLEVR~\cite{johnson2017CLEVRDiagnosticDataset} dataset through the addition of structured natural language explanations for each question-image pair. An example for our proposed CLEVR-X dataset is shown in Fig.~\ref{fig:teaser-dataset-comparison}.
The synthetic nature of the CLEVR-X dataset results in several advantages over datasets that use human explanations.
Since the explanations are synthetically constructed from the underlying scene graph, the explanations are \textit{correct} and do not require auxiliary prior knowledge. The synthetic textual explanations do not suffer from errors that get introduced with human explanations. Nevertheless, the explanations in the CLEVR-X dataset are human parsable as demonstrated in the human user study that we conducted. Furthermore, the explanations contain all the information that is necessary to answer a given question about an image without seeing the image. This means that the explanations are \textit{complete} with respect to the question about the image.

The CLEVR-X dataset allows for detailed diagnostics of natural language explanation generation methods in the context of VQA\@. For instance, it contains a wider range of question types than other related datasets. We provide baseline performances on the CLEVR-X dataset using recent frameworks for natural language explanations in the context of VQA\@. Those frameworks are jointly trained to answer the question and provide a textual explanation. Since the question family, question complexity (number of reasoning steps required), and the answer type (binary, counting, attributes) is known for each question and answer, the results can be analyzed and split according to these groups. In particular, the challenging counting problem~\cite{Trott2018InterpretableCF}, which is not well-represented in the VQA-X dataset, can be studied in detail on CLEVR-X.
Furthermore, our dataset contains multiple ground-truth explanations for each image-question pair. These capture a large portion of the space of correct explanations which allows for a thorough analysis of the influence of the number of ground-truth explanations used on the evaluation metrics.
Our approach of constructing textual explanations from a scene graph yields a great resource which could be extended to other datasets that are based on scene graphs, such as the CLEVR-CoGenT dataset.

To summarize, we make the following four contributions:
\begin{enumerate*}[label=(\arabic*)]
    \item We introduce the CLEVR-X dataset with natural language explanations for Visual Question Answering; 
    \item We confirm that the CLEVR-X dataset consists of correct explanations that contain sufficient relevant information to answer a posed question by conducting a user study;
    \item We provide baseline performances with two state-of-the-art methods that were proposed for generating textual explanations in the context of VQA;
    \item We use the CLEVR-X dataset for a detailed analysis of the explanation generation performance for different subsets of the dataset and to better understand the metrics used for evaluation.
\end{enumerate*}

\section{Related work}
In this section, we discuss several themes in the literature that relate to our work, namely \textit{Visual Question Answering}, \textit{Natural language explanations (for vision-language tasks)}, and the \textit{CLEVR dataset}. 

\myparagraph{Visual Question Answering (VQA).} The VQA~\cite{antol2015vqa} task has been addressed by several works that apply attention mechanisms to text and image features~\cite{yang2016stacked,xu2016ask,zhu2016visual7w,shih2016look,fukui2016MultimodalCompactBilinear}.
However, recent works observed that the question-answer bias in common VQA datasets can be exploited in order to answer questions without leveraging any visual information~\cite{agrawal2016analyzing,agrawal2018don,johnson2017CLEVRDiagnosticDataset,zhang2016yin}.
This has been further investigated in more controlled dataset settings, such as the CLEVR~\cite{johnson2017CLEVRDiagnosticDataset}, VQA-CP~\cite{agrawal2018don}, and GQA~\cite{hudson2019GQANewDataset} datasets. In addition to a controlled dataset setting, our proposed CLEVR-X dataset contains natural language explanations that enable a more detailed analysis of the reasoning in the context of VQA\@.

\myparagraph{Natural language explanations.}
Decisions made by neural networks can be visually explained with visual attribution that is determined by introspecting trained networks and their features~\cite{simonyan2013deep,zeiler2014visualizing,selvaraju2019GradCAMVisual,bach2015pixel,zhang2018top}, by using input perturbations~\cite{petsiuk2018rise,fong2019understanding,fong_iccv_2017}, or by training a probabilistic feature attribution model along with a task-specific CNN~\cite{kim2021keep}. Complementary to visual explanations methods that tend to not help users distinguish between correct and incorrect predictions~\cite{kim2021hive}, natural language explanations have been investigated for a variety of tasks, such as fine-grained visual object classification~\cite{hendricks2018GroundingVisualExplanations,hendricks2018Generatingcounterfactualexplanations}, or self-driving car models~\cite{kim2018textual}. The requirement to ground language explanations in the input image can prevent shortcuts, such as relying on dataset statistics or referring to instance attributes that are not present in the image.
For a comprehensive overview of research on explainability and interpretability, we refer to recent surveys~\cite{Barredo_Arrieta_XAI_2020,brundage2020toward,gilpin2018explaining}.

\myparagraph{Natural language explanations for vision-language tasks.}
Multiple datasets for natural language explanations in the context of vision-language tasks have been proposed, such as the VQA-X~\cite{hukpark2018MultimodalExplanationsJustifying}, VQA-E~\cite{li2018vqa}, and e-SNLI-VE datasets~\cite{kayser2021vil}.   VQA-X~\cite{hukpark2018MultimodalExplanationsJustifying} augments a small subset of the VQA~v2~\cite{goyal2017MakingVVQA} dataset for the Visual Question Answering task with human explanations.
Similarly, the VQA-E dataset~\cite{li2018vqa} extends the VQA~v2 dataset by sourcing explanations from image captions. However, the VQA-E explanations resemble image descriptions and do not provide satisfactory justifications whenever prior knowledge is required~\cite{li2018vqa}. The e-SNLI-VE~\cite{kayser2021vil,Do2020eSNLIVE20CV} dataset combines human explanations from e-SNLI~\cite{Camburu2018eSNLINL} and the image-sentence pairs for the Visual Entailment task from SNLI-VE~\cite{Xie2019VisualEA}.
In contrast to the VQA-E, VQA-X, and e-SNLI-VE datasets which consist of human explanations or image captions, our proposed dataset contains systematically constructed explanations derived from the associated scene graphs.
Recently, several works have aimed at generating natural language explanations for vision-language tasks~\cite{hukpark2018MultimodalExplanationsJustifying,wu2019FaithfulMultimodalExplanation,wu2020ImprovingVQAits,Marasovi2020NaturalLR,patro2020robust,kayser2021vil}.
In particular, we use the PJ-X~\cite{hukpark2018MultimodalExplanationsJustifying} and FM~\cite{wu2019FaithfulMultimodalExplanation} frameworks to obtain baseline results on our proposed CLEVR-X dataset.

\myparagraph{The CLEVR dataset.}
The CLEVR dataset~\cite{johnson2017CLEVRDiagnosticDataset} was proposed as a diagnostic dataset to inspect the visual reasoning of VQA models.
Multiple frameworks have been proposed to address the CLEVR task~\cite{hudson2018CompositionalAttentionNetworks,perezFiLMVisualReasoning,hudson2019learning,johnson2017inferring,Suarez2018DDRprogAC,Shi2019ExplainableAE}. 
To add explainability, the XNM model~\cite{Shi2019ExplainableAE} adopts the scene graph as an inductive bias which enables the visualization of the reasoning based on the attention on the nodes of the graph.
There have been numerous dataset extensions for the CLEVR dataset, for instance to measure the generalization capabilities of models pre-trained on CLEVR (CLOSURE~\cite{bahdanau2019closure}), to evaluate object detection and segmentation (CLEVR-Ref+~\cite{liu2019clevr}), or to benchmark visual dialog models (CLEVR dialog~\cite{kottur2019clevr}).
The Compositional Reasoning Under Uncertainty (CURI) benchmark uses the CLEVR renderer to construct a test bed for compositional and relational learning under uncertainty~\cite{vedantam2021curi}. \cite{Holzinger2021KANDINSKYPatternsA} provide an extensive survey of further experimental diagnostic benchmarks for analyzing explainable machine learning frameworks along with proposing the KandinskyPATTERNS benchmark that contains synthetic images with simple 2-dimensional objects. It can be used for testing the quality of explanations and concept learning.
Additionally,~\cite{arras2020ground} proposed the CLEVR-XAI-simple and CLEVR-XAI-complex datasets which provide ground-truth segmentation information for heatmap-based visual explanations.
Our CLEVR-X augments the existing CLEVR dataset with explanations, but in contrast to (heatmap-based) visual explanations, we focus on natural language explanations.

\section{The CLEVR-X dataset}\label{sec:dataclevrx}

In this section, we introduce the CLEVR-X dataset that consists of natural language explanations in the context of VQA\@. The CLEVR-X dataset extends the CLEVR dataset with 3.6 million natural language explanations for 850k question-image pairs.
In Section~\ref{sec:data-clevr}, we briefly describe the CLEVR dataset, which forms the base for our proposed dataset. Next, we present an overview of the CLEVR-X dataset by describing how the natural language explanations were obtained in Section~\ref{sec:data-generation}, and by providing a comprehensive analysis of the CLEVR-X dataset in Section~\ref{sec:data-analysis}.
Finally, in Section~\ref{sec:data-userstudy}, we present results for a user study on the CLEVR-X dataset.

\subsection{The  CLEVR dataset}\label{sec:data-clevr}
The CLEVR dataset consists of images with corresponding full scene graph annotations which contain information about all objects in a given scene (as nodes in the graph) along with spatial relationships for all object pairs. The synthetic images in the CLEVR dataset contain three to ten (at least partially visible) objects in each scene, where each object has the four distinct properties \texttt{size}, \texttt{color}, \texttt{material}, and \texttt{shape}.
There are three shapes (\texttt{box}, \texttt{sphere}, \texttt{cylinder}), eight colors (\texttt{gray}, \texttt{red}, \texttt{blue}, \texttt{green}, \texttt{brown}, \texttt{purple}, \texttt{cyan}, \texttt{yellow}), two sizes (\texttt{large}, \texttt{small}), and two materials (\texttt{rubber}, \texttt{metallic}). This allows for up to 96 different combinations of properties.

There are a total of 90 different question families in the dataset which are grouped into 9 different question types. Each type contains questions from between 5 and 28 question families.
In the following, we describe the 9 question types in more detail.

\myparagraph{\emph{Hop} questions:}
The \emph{zero hop}, \emph{one hop}, \emph{two hop}, and \emph{three hop} question types contain up to three relational reasoning steps, e.g.\ \enquote{What color is the cube \underline{to the left of} the ball?} is a \emph{one hop} question.

\myparagraph{\emph{Compare} and \emph{relate} questions:}
The \emph{compare integer}, \emph{same relate}, and \emph{comparison} question types require the understanding and comparison of multiple objects in a scene.
Questions of the \emph{compare integer} type compare counts corresponding to two independent clauses (e.g.\ \enquote{Are there \underline{more} cubes \underline{than} red balls?}).
\emph{Same relate} questions reason about objects that have the same attribute as another previously specified object (e.g.\ \enquote{What is the color of the cube that has \underline{the same size} as the ball?}).
In contrast, \emph{comparison} question types compare the attributes of two objects (e.g.\ \enquote{\underline{Is the color} of the cube \underline{the same} as the ball?}).

\myparagraph{\emph{Single and/or} questions:}
\emph{Single or} questions identify objects that satisfy an exclusive disjunction condition (e.g.\ \enquote{How many objects are \underline{either} red \underline{or} blue?}).
Similarly, \emph{single and} questions apply multiple relations and filters to find an object that satisfies all conditions (e.g.\ \enquote{How many objects are red \underline{and} to the left of the cube.}).\\

\rev{Each CLEVR question can be represented by a corresponding functional program and its natural language realization. 
A functional program is composed of basic functions that resemble elementary visual reasoning operations, such as \emph{filtering} \rev{objects} by \rev{one or more} properties, spatially \emph{relating} objects to each other, or \emph{querying} object properties.
Furthermore, logical operations \rev{like} \emph{and} and \emph{or}, as well as counting operations like \emph{count}, \emph{less}, \emph{more}, and \emph{equal} are used to build complex questions.
Executing the functional program associated with the question against the scene graph yields the correct answer to the question.}
We can distinguish between three different answer types: Binary answers (\texttt{yes} or \texttt{no}), counting answers (integers from \texttt{0} to \texttt{10}), and attribute answers (any of the possible values of \texttt{shape}, \texttt{color}, \texttt{size}, or \texttt{material}).

\subsection{Dataset generation}%
\label{sec:data-generation}
\begin{figure}[t]
     \centering
        \centering
        \includegraphics[width=\linewidth,trim=3mm 85mm 7mm 15mm,clip]{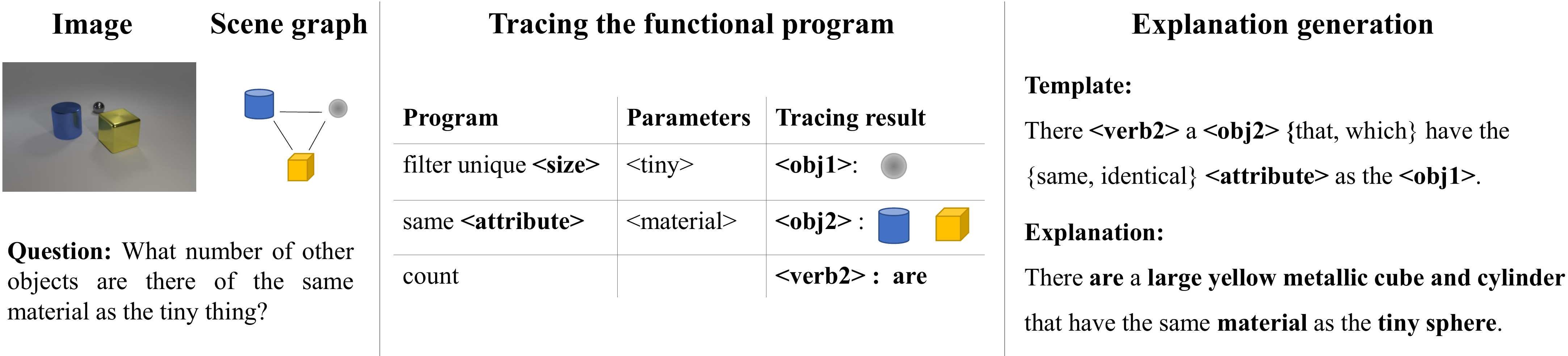}
        \caption{%
        CLEVR-X dataset generation: Generating a natural language explanation for a sample from the CLEVR dataset. Based on the question, the functional program for answering the question is executed on the scene graph and traced. A language template is used to cast the gathered information into a natural language explanation.
        }%
        \label{fig:generation-and-examples}
\end{figure}

Here, we describe the process for generating natural language explanations for the CLEVR-X dataset.
\rev{In contrast to image captions, the CLEVR-X explanations only describe image elements that are relevant to a specific input question.}
The explanation generation process for a given question-image pair is illustrated in Fig.~\ref{fig:generation-and-examples}. It consists of three steps: Tracing the functional program, relevance filtering (not shown in the figure), and explanation generation. In the following, we will describe those steps in detail.

\myparagraph{Tracing the functional program.}
Given a question-image pair from the CLEVR dataset, we trace the execution of the functional program (that corresponds to the question) on the scene graph (which is associated with the image). The generation of the CLEVR dataset uses the same step to obtain a question-answer pair. When executing the basic functions that comprise the functional program, we record their outputs in order to collect all the information required for explaining a ground-truth answer.

\rev{In particular, we trace} the \emph{filter}, \emph{relate} and \emph{same-property} functions \rev{and record the returned objects and their properties, such as, for instance, \texttt{shape} or \texttt{size}.}
As a result, the tracing omits objects \rev{in the scene} that are not relevant for the question.
\rev{As we are aiming for complete explanations for all question types, each explanation has to mention all the objects that were needed to answer the question, i.e.\ all the evidence that was obtained during tracing.
For example, for \emph{counting} questions, all objects that match the \rev{\emph{filter} function} preceding the \emph{counting} step are recorded during tracing.
For \emph{and} questions, we merge the tracing results of the preceding \rev{functions} which results in short and readable explanations.}
\rev{In summary, the tracing} produces a \emph{complete} \rev{and \emph{correct}} understanding of the objects and relevant properties which contributed to an answer.

\myparagraph{Relevance filtering.}
To keep the explanation at a reasonable length, we filter the object attributes that are mentioned in the explanation according to their relevance.
For example, the \texttt{color} of an object is not relevant for a given question that asks about the \texttt{material} of said object.
\rev{We deem all properties that were listed in the question to be relevant.}
This makes it easier to recognize the same referenced object in both the question and explanation.
As the \texttt{shape} property also serves as a noun in CLEVR, our explanations always mention the \texttt{shape} to avoid using generic shape descriptions like \enquote{object} or \enquote{thing}.
We distinguish between objects which are used to build the question (\eg{} \enquote{[\ldots] that is left of the \textit{cube}?}) and those that are the subject of the posed question (\eg{} \enquote{What color is the \textit{sphere} that is left of the cube?}).
\rev{For the former, we do not mention any additional properties, and for the latter}, we mention the queried property (e.g.\ \texttt{color}) for question types yielding attribute answers.

\myparagraph{Explanation generation.} To obtain the final natural language explanations, each question type is equipped with one or more natural language templates with variations in terms of the wording used.
Each template contains placeholders which are filled with the output of the previous steps, i.e.\ the tracing of the functional program and subsequent filtering for relevance.
As mentioned above, our explanations use the same property descriptions that appeared in the question. 
\rev{This is done to ensure that the wording of the explanation is consistent with the given question, e.g.\ for the question \enquote{Is there a \underline{small} object?} we generate the explanation \enquote{Yes there is a \underline{small} cube.}%
\footnote{%
The explanation could have used the synonym \enquote{box} instead of \enquote{cube}. In contrast, \enquote{tiny} and \enquote{small} are also synonyms in CLEVR, but the explanation would not have been consistent with the question which used \enquote{small}.}%
}.
We randomly sample synonyms for describing the properties of objects that do not appear in the question.
If multiple objects are mentioned in the explanation, we randomize their order.
\rev{If the tracing step returned an empty set, e.g.\ if no object exists that matches the given filtering function for an \emph{existence} or \emph{counting} question, we state that no relevant object is contained in the scene (e.g.\ \enquote{There is no red cube.}).}

In order to decrease the overall sentence length and to increase the readability, we aggregate repetitive descriptions (e.g.\ \enquote{There is a \underline{red cube} and a \underline{red cube}}) using numerals (e.g.\ \enquote{There are \underline{two} red cubes.}).
In addition, if \rev{a function of} the functional program merely restricts the output set of a preceding \rev{function}, we only mention the outputs of the later \rev{function}.
For instance, if a \texttt{same-color} \rev{function} yields a large and a small cube, and a \rev{subsequent} \texttt{filter-large} \rev{function} restricts th\rev{e output} to only the large cube, we do not mention the output of \texttt{same-color}, as the output of the following \texttt{filter-large} causes natural language redundancies%
\footnote{\rev{E.g.\ for the question: \enquote{How many large objects have the same color as the cube?}, we do not generate the explanation \enquote{There are a small and a large cube that have the same color as the red cylinder of which only the large cube is large.} but instead only write \enquote{There is a large cube that has the same color as the red cylinder.}}}%
.

\rev{The selection of different language templates, random sampling of synonyms and randomization of the object order (if possible) results in multiple different explanations.} 
We uniformly sample up to 10 different explanations \rev{per question }for our dataset.

\myparagraph{Dataset split.}
We provide explanations for the CLEVR training and validation sets, skipping only a negligible subset (less than $0.04\permil$) of questions due to malformed question programs from the CLEVR dataset\rev{, e.g.\ due to disjoint parts of their abstract syntax trees. In total, this affected 25 CLEVR training and 4 validation questions.}

As the scene graphs and question functional programs are not publicly available for the \rev{CLEVR} test set, we use the original CLEVR validation subset as the CLEVR-X test set. 20\% of the CLEVR training set serve as the CLEVR-X validation set. We perform this split on the image-level to avoid any overlap between images in the CLEVR-X training and validation sets.
Furthermore, we verified that the relative proportion of samples from each question and answer type in the CLEVR-X training and validation sets is similar, such that there are no biases towards specific question or answer types. 

Code for generating the CLEVR-X dataset and the dataset itself are publicly available at \url{https://github.com/ExplainableML/CLEVR-X}.

\subsection{Dataset analysis}\label{sec:data-analysis}
\setlength{\tabcolsep}{3pt}
\renewcommand{\arraystretch}{1.2}
\begin{table}[t]
    \centering
    \caption{%
    Statistics of the CLEVR-X dataset compared to the VQA-X, and e-SNLI-VE datasets. We show the total number of images, questions, and explanations, vocabulary size, 
  and the average number of explanations per question, the average number of words per explanation, and the average number of words per question.
    Not all subset values add up to the Total values since some subsets have overlaps (e.g.\ for the vocabulary).
    }%
    \label{tab:new_dataset_statistics}
    \resizebox{\linewidth}{!}{%
        \begin{tabular}{llccccccc}
            \toprule
            \multirow{2}{*}{Dataset}   & \multirow{2}{*}{Subset} & \multicolumn{4}{c}{Total \#} & \multicolumn{3}{c}{Average \# }                                                                         \\
            \cmidrule(r){3-6}\cmidrule(r){7-9}
                                       &                        & Images                       & Questions                       & Explanations & Vocabulary & Explanations & Expl. Words & Quest. Words \\
            \midrule
            \multirow{4}{*}{VQA-X}     & Train                  & 24,876                       & 29,549                          & 31,536       & 9,423      & 1.07        & 10.55       & 7.50         \\[-0.35ex]
                                       & Val                    & 1,431                        & 1,459                           & 4,377        & 3,373      & 3.00        & 10.88       & 7.56         \\[-0.35ex]
                                       & Test                   & 1,921                        & 1,921                           & 5,904        & 3,703      & 3.07        & 10.93       & 7.31         \\
                                       & Total                  & 28,180                       & 32,886                          & 41,817       & 10,315     & 1.48        & 10.64       & 7.49         \\[0.6ex] 

            \multirow{4}{*}{e-SNLI-VE} & Train                  & 29,779                       & 401,672                         & 401,672      & 36,778     & 1.00        & 13.62       & 8.23         \\[-0.35ex]
                                       & Val                    & 1,000                        & 14,339                          & 14,339       & 8,311      & 1.00        & 14.67       & 8.10         \\[-0.35ex]
                                       & Test                   & 998                          & 14,712                          & 14,712       & 8,334      & 1.00        & 14.59       & 8.20         \\
                                       & Total                  & 31,777                       & 430,723                         & 430,723      & 38,208     & 1.00        & 13.69       & 8.23         \\  
            \midrule
            \multirow{4}{*}{CLEVR-X}   & Train                  & 56,000                       & 559,969                         & 2,401,275    & 96         & 4.29         & 21.52       & 21.61        \\[-0.35ex]
                                       & Val                    & 14,000                       & 139,995                         & 599,711      & 96         & 4.28         & 21.54       & 21.62        \\[-0.35ex]
                                       & Test                   & 15,000                       & 149,984                         & 644,151      & 96         & 4.29         & 21.54       & 21.62        \\
                                       & Total                  & 85,000                       & 849,948                         & 3,645,137    & 96         & 4.29         & 21.53       & 21.61        \\  
            \bottomrule
        \end{tabular}
    }
\end{table}

\begin{figure}[t]
     \centering
     \begin{subfigure}[t]{0.495\textwidth}
        \centering
        \includegraphics[width=\linewidth,trim=3mm 3mm 3mm 4mm, clip]{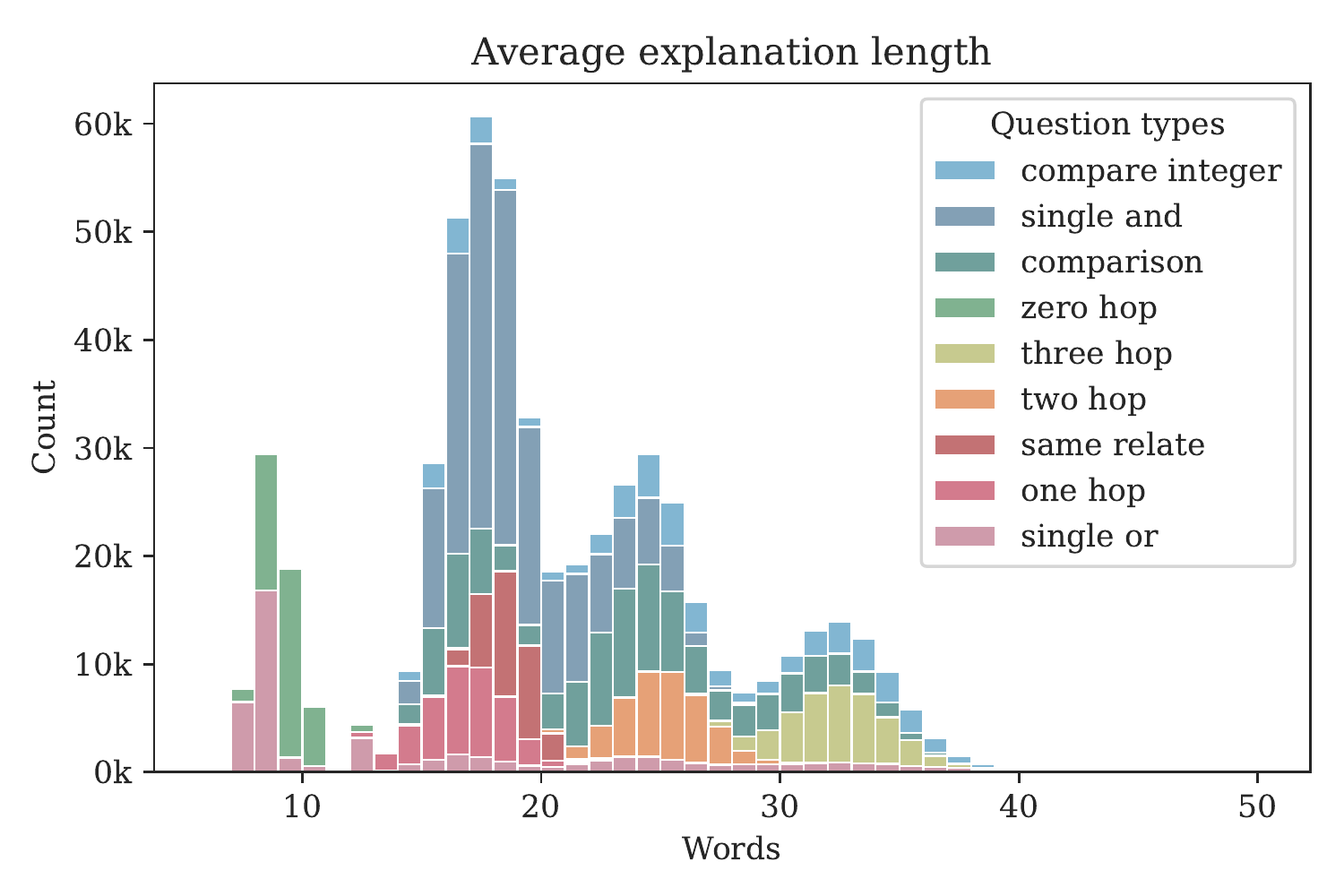}
     \end{subfigure}
     \hfill
     \begin{subfigure}[t]{0.495\textwidth}
        \centering
        \includegraphics[width=\linewidth,trim=3mm 3mm 3mm 4mm, clip]{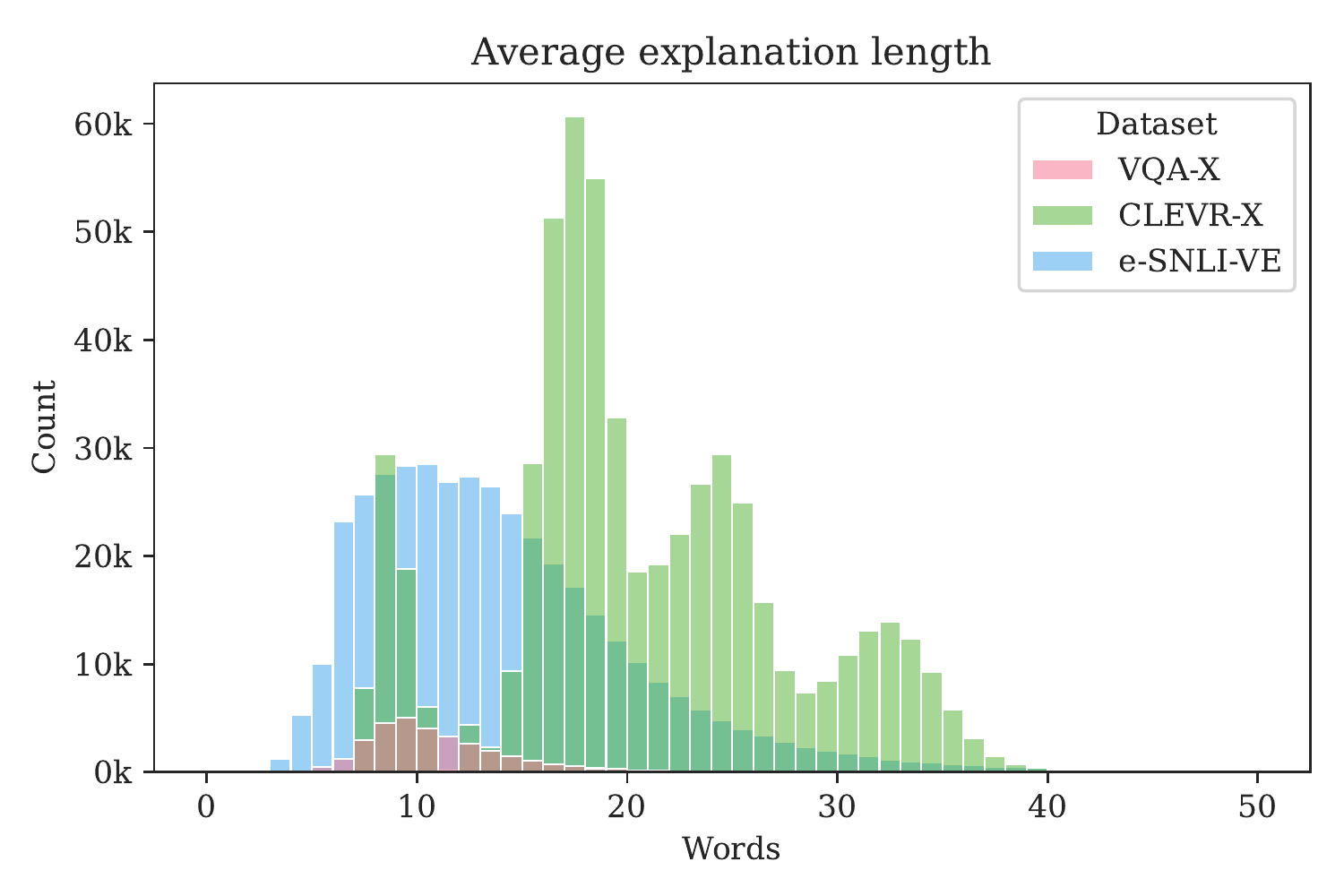}
     \end{subfigure}
     \vspace{-2ex}
    \caption{%
        Stacked histogram of the average explanation lengths measured in words for the 9 question types for the CLEVR-X training set (left).
        Explanation length distribution for the \clevrx{}, \vqax{}, and \esnlive{} \rev{training sets} (right).
        The long tail of the \esnlive{} distribution (125 words) was cropped out \rev{for better} readability.
    }%
    \label{fig:dataset-statisics}
    \vspace{-0.6em}
\end{figure}

We compare the CLEVR-X dataset to the related VQA-X and e-SNLI-VE datasets in Table~\ref{tab:new_dataset_statistics}. Similar to CLEVR-X, VQA-X contains natural language explanations for the VQA task. However, different to the natural images and human explanations in VQA-X, CLEVR-X consists of synthetic images and explanations.
The \esnlive{} dataset provides explanations for the visual entailment (VE) task. VE consists of classifying an input image-hypothesis pair into entailment / neutral / contradiction categories.

The \clevrx{} dataset is significantly larger than the \vqax{} and \esnlive{} datasets in terms of the number of images, questions, and explanations.
In contrast to the two other datasets, \clevrx{} provides (on average) multiple explanations for each question-image pair in the train set.
Additionally, the average number of words per explanation is also higher. Since the explanations are built to explain each component mentioned in the question, long questions require longer explanations than short questions. Nevertheless, by design, there are no unnecessary redundancies. The explanation length in \clevrx is very strongly correlated with the length of the corresponding question (Spearman's correlation coefficient between the number of words in the explanations and questions is $0.89$). 

Figure~\ref{fig:dataset-statisics} (left) shows the explanation length distribution in the CLEVR-X dataset for the 9 question types.
The shortest explanation consists of 7 words, and the longest one has 53 words. On average, the explanations contain \rev{21.53} words. In Fig.~\ref{fig:dataset-statisics} (right) and Table~\ref{tab:new_dataset_statistics}, we can observe that explanations in CLEVR-X tend to be longer than the explanations in the VQA-X dataset. Furthermore, \vqax has significantly fewer samples overall than the CLEVR-X dataset. The e-SNLI-VE dataset also contains longer explanations (that are up to 125 words long), but the CLEVR-X dataset is significantly larger than the e-SNLI-VE dataset.
However, due to the synthetic nature and limited domain of CLEVR, the vocabulary of CLEVR-X is very small with only 96 different words.
Unfortunately, \vqax{} and \esnlive{} contain spelling errors, resulting in multiple versions of the same words.
Models trained on CLEVR-X circumvent those aforementioned challenges and can purely focus on visual reasoning and explanations for the same.
Therefore, Natural Language Generation (NLG) metrics applied to CLEVR-X indeed capture the factual correctness and completeness of an explanation.

\subsection{User study on explanation completeness and relevance}\label{sec:data-userstudy}

In this section, we describe our user study for evaluating the completeness and relevance of the generated ground-truth explanations in the \clevrx{} dataset.
\rev{%
We wanted to verify whether humans are successfully able to parse the synthetically generated textual explanations and to select complete and relevant explanations.
While this is obvious for easier explanations like \enquote{There is a blue sphere.}, it is less trivial for more complex explanations such as \enquote{There are two red cylinders in front of the green cube that is to the right of the tiny ball.}
Thus, strong human performance in the user study indicates that the sentences are parsable by humans.
}

\begin{figure}[t]
    \centering
    \begin{subfigure}[t]{0.48\textwidth}
        \centering
        \includegraphics[height=8.3cm]{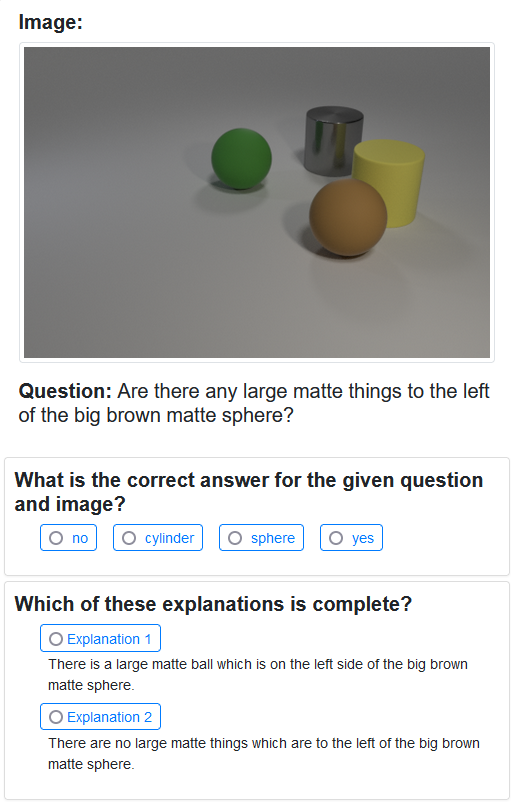}%
        \label{fig:user-study-example-completeness}
    \end{subfigure}
    \hfill
    \begin{subfigure}[t]{0.48\textwidth}
        \centering
        \includegraphics[height=8.3cm]{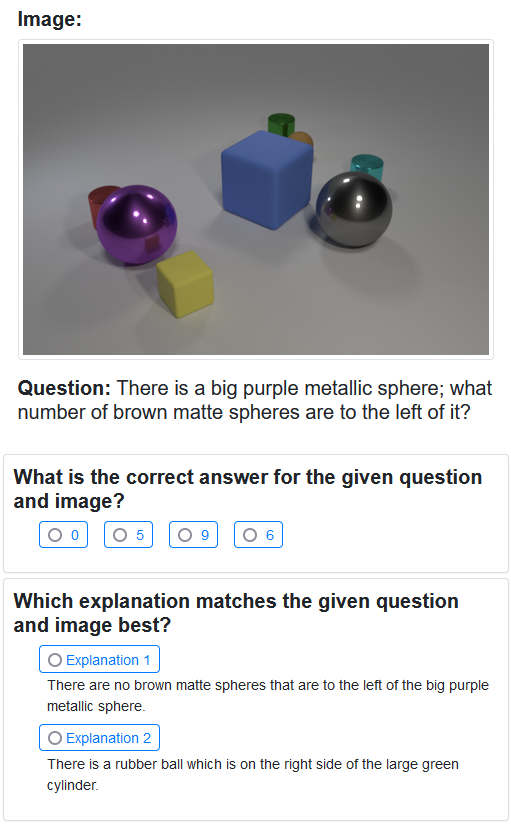}%
        \label{fig:user-study-example-relevance}
    \end{subfigure}%
    \caption{Two examples from our user study to evaluate the completeness (left) and relevance (right) of natural language explanations in the CLEVR-X dataset.}\label{fig:user-study-example}
\end{figure}

We performed our user study using Amazon Mechanical Turk (MTurk). It consisted of two types of Human Intelligence Tasks (HITs).
Each HIT was made up of
\begin{enumerate*}[label=(\arabic*)]
\item An explanation of the task;
\item A non-trivial example, where the correct answers are already selected;
\item A CAPTCHA~\cite{Ahn2003CAPTCHAUH} to verify that the user is human;
\item The problem definition consisting of a question and an image;
\item A \rev{user qualification step}, for which the user has to correctly answer a question about an image. \rev{This ensures that the user is able to answer the question in the first place, a necessary condition to participate in our user study};
\item Two explanations from which the user needs to choose one.
\end{enumerate*}
Example screenshots of the user interface for the user study are shown in Fig.~\ref{fig:user-study-example}.

For the two different HIT types, we randomly sampled 100 explanations from each of the 9 question types, resulting in a total of 1800 samples for the completeness and relevance tasks. 
For each task sample, we requested 3 different MTurk workers based in the US (with high acceptance rate of $>95\%$ and over 5000 accepted HITs).
A total of 78 workers participated in the completeness HITs. They took on average 144.83 seconds per HIT\@.
The relevance task was carried out by 101 workers which took on average 120.46 seconds per HIT\@.
In total, 134 people participated in our user study.
In the following, we describe our findings regarding the completeness and relevance of the CLEVR-X explanations in more detail.

\myparagraph{Explanation completeness.}
In the first part of the user study, we evaluated whether human users are able to determine if the ground-truth explanations in the \clevrx{} dataset are complete (and also correct).
We presented the MTurk workers with an image, a question, and two explanations. 
As can be seen in Fig.~\ref{fig:user-study-example} (left), a user had to first select the correct answer (\textit{yes}) before deciding which of the two given explanations was complete.
By design, one of the explanations presented to the user was the complete one from the CLEVR-X dataset and the other one was a modified version for which at least one necessary object had been removed. As simply deleting an object from a textual explanation could lead to grammar errors, we re-generated the explanations after removing objects from the tracing results. This resulted in incomplete, albeit grammatically correct, explanations.

To evaluate the ability to determine the completeness of explanations, we measured the accuracy of selecting the complete explanation.
The human participants obtained an average accuracy of \rev{92.19}\%, confirming that  complete explanations which mention all objects necessary to answer a given question were preferred over incomplete ones. 
The performance was weaker for complex question types, such as \textit{compare-integer} and \textit{comparison} with accuracies of only 77.00\% and 83.67\% respectively, compared to the easier \textit{zero-hop} and \textit{one-hop} questions with accuracies of 100\% and 98.00\% respectively. 

\begin{figure}[t]
    \centering
    \includegraphics[width=\linewidth,trim=4mm 4mm 4mm 4mm,clip]{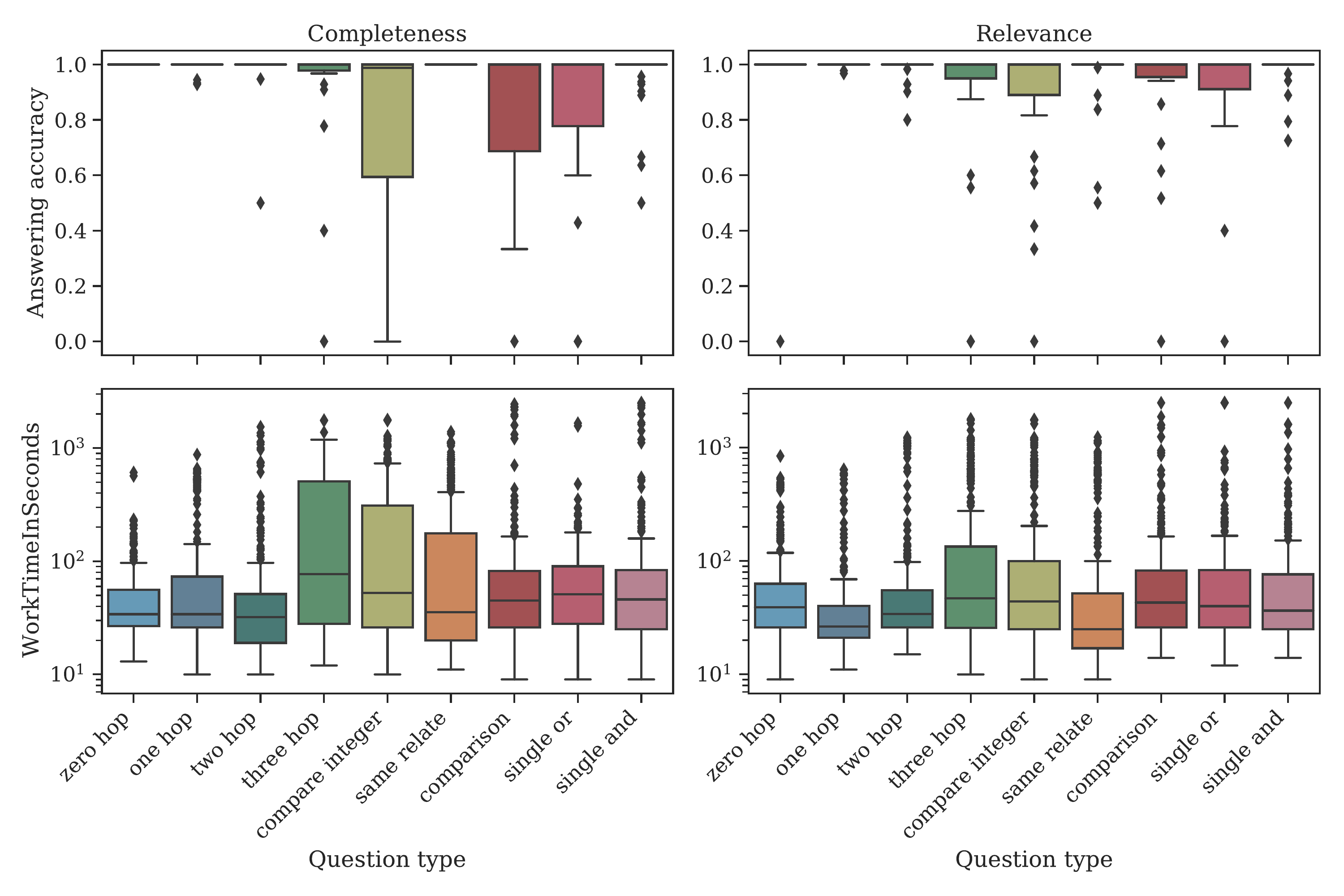}
    \caption{%
    \rev{Average answering accuracies for each worker (top) and average work time (bottom) for the user study (left: completeness, right: relevance). The boxes indicate the mean as well as lower and upper quartiles, the lines extend 1.5 interquartile ranges of the lower and upper quartile. All other values are plotted as diamonds.
    }
    }%
    \label{fig:user-study-worker-answer-accuracies-and-work-time}%
\end{figure}

Additionally, there were huge variations in performance across different participants of the completeness study (Fig.~\ref{fig:user-study-worker-answer-accuracies-and-work-time}~(top left)), with the majority performing very well ($>$97\% answering accuracy) for most question types. For the \emph{compare-integer}, \emph{comparison} and \emph{single or} question types, some workers exhibited a much weaker performance with answering accuracies as low as $0\%$.
The average turnaround time shown in Fig.~\ref{fig:user-study-worker-answer-accuracies-and-work-time}~(bottom left) confirms that easier question types required less time to be solved than more complex question types, such as \emph{three hop} and \emph{compare integer} questions. Similar to the performance, the work time varied greatly between different users.

\begin{table}[t]
    \centering
        \caption{Results for the user study evaluating the accuracy for the completeness and relevance tasks for the 9 question types in the CLEVR-X dataset.}%
    \resizebox{\textwidth}{!}{%
    \begin{tabulary}{1.125\textwidth}{p{2cm} C C C C C C C C C C}
    \toprule
         & Zero hop & One hop & Two hop & Three hop & Same relate & Comparison & Compare integer & Single or & Single and & \multirow{2}{*}{All}\\
    \midrule
        Completeness        & 100.00  &   98.00  &  98.67  &   94.00   &   100.00    &    83.67   &    77.00        &  84.00  & 94.33 & \rev{92.19}\\
        Relevance           &  99.67  &   99.00   &  95.67  &   89.00   &   95.67     &    87.33   &    83.67        &  90.67  & 92.00 & 92.52\\
    \bottomrule
    \end{tabulary}
    }%
    \label{tab:user-study-results}%
 \end{table}

\myparagraph{Explanation relevance.}
In the second part of our user study, we analyzed if humans are able to identify explanations which are relevant for a given image. For a given question-image pair, the users had to first select the correct answer. Furthermore, they were provided with a correct explanation and another randomly chosen explanation from the same question family (that did not match the image). The task consisted of selecting the correct explanation that matched the image and question content. Explanation~1 in the example user interface shown in Fig.~\ref{fig:user-study-example} (right) was the relevant one, since Explanation 2 does not match the question and image. 

The participants of our user study were able to determine which explanation matched the given question-image example with an average accuracy of 92.52\%. Again, the performance for complex question types was weaker than for easier questions. The difficulty of the question influences the accuracy of detecting the relevant explanation, since this task first requires understanding the question. Furthermore, complex questions tend to be correlated with complex scenes that contain many objects which makes the user's task more challenging. The accuracy for \textit{three-hop} questions was 89.00\% compared to 99.67\% for \textit{zero-hop} questions. For \textit{compare-integer} and \textit{comparison} questions, the users obtained accuracies of 83.67\% and 87.33\% respectively, which is significantly lower than the overall average accuracy.

We analyzed the answering accuracy per worker in Fig.~\ref{fig:user-study-worker-answer-accuracies-and-work-time}~(top right).
The performance varies greatly between workers, with the majority performing very well ($>$90\% answering accuracy) for most question types.
Some workers showed much weaker performance with answering accuracies as low as $0\%$ (e.g.\ for \emph{compare-integer} and \emph{single or} questions).
Furthermore, the distribution of work time for the relevance task is shown in Fig.~\ref{fig:user-study-worker-answer-accuracies-and-work-time}~(bottom right). The turnaround times for each worker exhibit greater variation on the completeness task (bottom left) compared to the relevance task (bottom right).
This might be due to the nature of the different tasks. For the completeness task, the users need to check if the explanation contains all the elements that are necessary to answer the given question. The relevance task, on the other hand, can be solved by detecting a single non-relevant object to discard the wrong explanation.

Our user study confirmed that humans are able to parse the synthetically generated natural language explanations in the CLEVR-X dataset. Furthermore, the results have shown that users prefer complete and relevant explanations in our dataset over corrupted samples.
\section{Experiments}
We describe the experimental setup for establishing baselines on our proposed CLEVR-X dataset in Section~\ref{sec:exp-setup}. 
In Section~\ref{sec:clevrx-results}, we present quantitative results on the CLEVR-X dataset. Additionally, we analyze the generated explanations for the CLEVR-X dataset in relation to the question and answer types in Section~\ref{sec:type-results}.
Furthermore, we study the behavior of the NL\rev{G} metrics when using different numbers of ground-truth explanations for testing in Section~\ref{sec:length-results}.
Finally, we present qualitative explanation generation results on the CLEVR-X dataset in Section~\ref{sec:quali-results}.

\subsection{Experimental setup}\label{sec:exp-setup}
In this section, we provide details about the datasets and models used to establish baselines for our CLEVR-X dataset and about their training details. Furthermore, we explain the metrics for evaluating the explanation generation performance.

\myparagraph{Datasets.}
In the following, we summarize the datasets that were used for our experiments. In addition to providing baseline results on CLEVR-X, we also report experimental results on the VQA-X and e-SNLI-VE datasets.
Details about our proposed \textbf{CLEVR-X} dataset can be found in Section~\ref{sec:dataclevrx}.
The \textbf{VQA-X} dataset~\cite{hukpark2018MultimodalExplanationsJustifying} is a subset of the VQA~v2 dataset with a single human-generated textual explanation per question-image pair in the training set and $3$ explanations for each sample in the validation and test sets.
The \textbf{e-SNLI-VE} dataset~\cite{Do2020eSNLIVE20CV,kayser2021vil} is a large-scale dataset with natural language explanations for the visual entailment task.

\myparagraph{Methods.}
We used multiple frameworks to provide baselines on our proposed CLEVR-X dataset.
For the \textbf{random words} baseline, we sample random word sequences of length $w$ for the answer and explanation words for each test sample. The full vocabulary corresponding to a given dataset is used as the sampling pool, and $w$ denotes the average number of words forming an answer and explanation in a given dataset.
For the \textbf{random~explanations} baseline, we randomly sample an answer-explanation pair from the training set and use this as the prediction. \rev{The explanations from this baseline are well-formed sentences. However, the answers and explanations most likely do not match the question or the image.}
For the random-words and random-explanations baselines, we report the NLG metrics for all samples in the test set (instead of only considering the correctly answered samples, since the random sampling of the answer does not influence the explanation).
The Pointing and Justification model \textbf{PJ-X}~\cite{hukpark2018MultimodalExplanationsJustifying} provides text-based post-hoc justifications for the VQA task. It combines a modified MCB~\cite{fukui2016MultimodalCompactBilinear} framework, pre-trained on the VQA v2 dataset, with a visual pointing and textual justification module.
The Faithful Multimodal (\textbf{FM}) model~\cite{wu2019FaithfulMultimodalExplanation} aims at grounding parts of generated explanations in the input image to provide explanations that are \textit{faithful} to the input image. It is based  on the Up-Down VQA model~\cite{anderson2018bottom}. In addition, FM contains an explanation module which enforces consistency between the predicted answer, explanation and the attention of the VQA model.
The implementations for the PJ-X and FM models are based on those provided by the authors of~\cite{kayser2021vil}.

\myparagraph{Implementation and training details.}
We extracted 14$\times$14$\times$1024 grid features for the images in the CLEVR-X dataset using a ResNet-101~\cite{he2016deep}, pre-trained on ImageNet~\cite{deng2009imagenet}. These grid features served as inputs to the FM~\cite{wu2019FaithfulMultimodalExplanation} and PJ-X~\cite{hukpark2018MultimodalExplanationsJustifying} frameworks. 
The CLEVR-X explanations are lower case and punctuation is removed from the sentences.
We selected the best model on the CLEVR-X validation set
based on the highest \rev{mean of the four NLG metrics, where explanations for incorrect answers were set to an empty string}. \rev{This metric accounts for the answering performance as well as for the explanation quality.}
The final models were 
evaluated on the CLEVR-X test set.
For PJ-X, our best model was trained for \rev{52} epochs, using the Adam optimizer~\cite{kingma2014adam} with a learning rate of 0.0002 and a batch size of \rev{256}.
We did not use gradient clipping for PJ-X.
Our strongest FM model was trained for \rev{30} epochs, using the Adam optimizer with a learning rate of 0.000\rev{2}, a batch size of 128, and gradient clipping of 0.1. All other hyperparameters were taken from~\cite{hukpark2018MultimodalExplanationsJustifying,wu2019FaithfulMultimodalExplanation}.

\myparagraph{Evaluation metrics.}
To evaluate the quality of the generated explanations, we use the standard natural language generation metrics BLEU~\cite{papineni2001BLEUMethodAutomatic}, METEOR~\cite{banerjee2005METEORAutomaticMetric}, ROUGE-L~\cite{lin2004ROUGEPackageAutomatic} and CIDEr~\cite{vedantam2015CIDErConsensusbased}.
By design, there is no correct explanation that can justify a wrong answer. 
We follow~\cite{kayser2021vil} and \rev{report the quality of the generated explanations for the subset of correctly answered questions}.

\subsection{Evaluating explanations generated by state-of-the-art methods}\label{sec:clevrx-results}
In this section, we present quantitative results for generating explanations for the CLEVR-X dataset (Table~\ref{tab:clevr-results}).
The random words baseline exhibits weak explanation performance for all NLG metrics on CLEVR-X.
Additionally, the random answering accuracy is very low at 3.6\%.
The results are similar on \vqax{} and \esnlive{}.
The random explanations baseline achieves stronger explanation results on all three datasets, but is still significantly worse than the trained models.
This confirms that, even with a medium-sized answer space (28 options) and a small vocabulary (96 words), it is not possible to achieve good scores on our dataset using a trivial approach.

We observed that the PJ-X model yields a significantly stronger performance on CLEVR-X in terms of the NL\rev{G} metrics for the generated explanations compared to the FM model, with METEOR scores of \rev{58.9} and \rev{52.5} for PJ-X and FM respectively. Across all explanation metrics, the scores on the VQA-X and e-SNLI-VE datasets are in a lower range than those on CLEVR-X. For PJ-X, we obtain a CIDEr score of \rev{639.8} on CLEVR-X and \rev{82.7} and \rev{72.5} on VQA-X and e-SNLI-VE\@. This can be attributed to the smaller vocabulary and longer sentences, which allow $n$-gram based metrics (e.g.\ BLEU) to match parts of sentences more easily.

\rev{In contrast to the explanation generation performance, the FM model is better at answering questions than PJ-X on CLEVR-X with an answering accuracy of 80.3\% for FM compared to 63.0\% for PJ-X.}
\rev{Compared to recent models tuned to the CLEVR task, the answering performances of PJ-X and FM do not seem very strong. However, the PJ-X backbone MCB~\cite{fukui2016MultimodalCompactBilinear} (which is crucial for the answering performance) preceded the publication of the CLEVR dataset.
A version of the MCB backbone (CNN+LSTM+MCB in the CLEVR publication~\cite{johnson2017CLEVRDiagnosticDataset}) achieved an answering accuracy of 51.4\% on CLEVR~\cite{johnson2017CLEVRDiagnosticDataset}, whereas PJ-X is able to correctly answer 63\% of the questions.
The strongest model discussed in the initial CLEVR publication (CNN+LSTM+SA in~\cite{johnson2017CLEVRDiagnosticDataset}) achieved an answering accuracy of 68.5\%.
}

{%
\setlength{\tabcolsep}{2pt}
\renewcommand{\arraystretch}{1.2} 
\begin{table}[t]
\centering
    \caption{Explanation generation results on the CLEVR-X, VQA-X, and e-SNLI-VE test sets using BLEU-4 (B4), METEOR (M), ROUGE-L (RL), CIDEr (C), and answer accuracy (\textit{Acc}). Higher is better for all reported metrics.
    For the random baselines, \textit{Acc} corresponds to $\nicefrac{100}{\text{\# answers}}$ for CLEVR-X and e-SNLI-VE, and to the VQA answer score for VQA-X. (Rnd.\ words: random words, Rnd.\ expl: Random explanations)%
    }%
    \label{tab:clevr-results}%
    \resizebox{\textwidth}{!}{%
    \begin{tabular}{l ccccc|ccccc|cccccc}
            \toprule
            \multirow{2}{3pt}{Model}                                & \multicolumn{5}{c|}{\textit{CLEVR-X}} & \multicolumn{5}{c|}{\textit{VQA-X}} & \multicolumn{5}{c}{\textit{e-SNLI-VE}}                                                                                     \\
                                                                    & B4                                    & M                                   & RL                                     & C     & Acc  & B4   & M    & RL   & C    & Acc  & B4  & M    & RL   & C    & Acc  \\
            \cmidrule(r){1-1}\cmidrule(r){2-6}\cmidrule(r){7-11}\cmidrule(r){12-16}
            Rnd.\ words                                      &   0.0                             & 8.4 & 11.4 & 5.9  & 3.6 & 0.0 & 1.2 & 0.7 & 0.1 & 0.1  & 0.0 & 0.3 & 0.0 & 0.0  & 33.3  \\ 
            Rnd.\ expl                                     &    10.9                           &  16.6                              & 35.3                                & 30.4 & 3.6 &  0.9 & 6.5 & 18.4 & 21.6 & 0.2 & 0.4 & 5.4 & 9.9 & 2.6 & 33.3  \\ 
            \midrule
            FM~\cite{wu2019FaithfulMultimodalExplanation}           &    78.8                             &          52.5                     &    85.8                           & 566.8   &  80.3 & 23.1 & 20.4 & 47.1 & 87.0 & 75.5 & 8.2 & 15.6 & 29.9 & 83.6 & 58.5 \\
            PJ-X~\cite{hukpark2018MultimodalExplanationsJustifying} &    87.4                                 &    58.9                        &  93.4                                 & 639.8  & 63.0 & 22.7 & 19.7 & 46.0 & 82.7 & 76.4 & 7.3 & 14.7 & 28.6 & 72.5 & 69.2 \\  
            \bottomrule
        \end{tabular}
    }
\end{table}
}

\subsection{Analyzing results on CLEVR-X by question and answer types}\label{sec:type-results}

\begin{figure}[t]
    \centering
    \begin{subfigure}[t]{0.6775\textwidth}
        \centering
        \includegraphics[width=\linewidth]{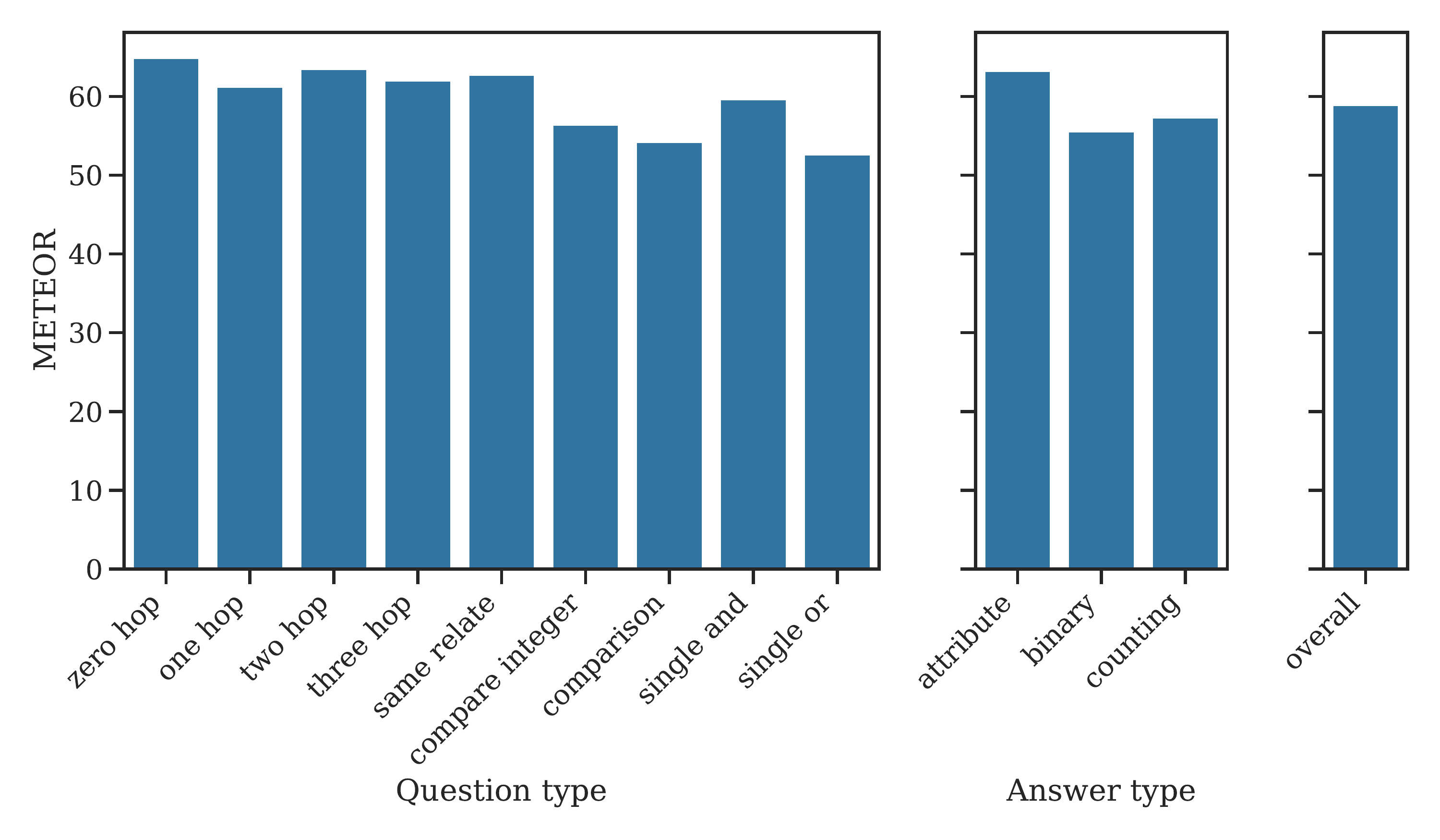}
    \end{subfigure}
    \hfill
    \begin{subfigure}[t]{0.2825\textwidth}
        \centering
    \includegraphics[width=\linewidth,trim=4mm 4mm 4mm 4mm,clip]{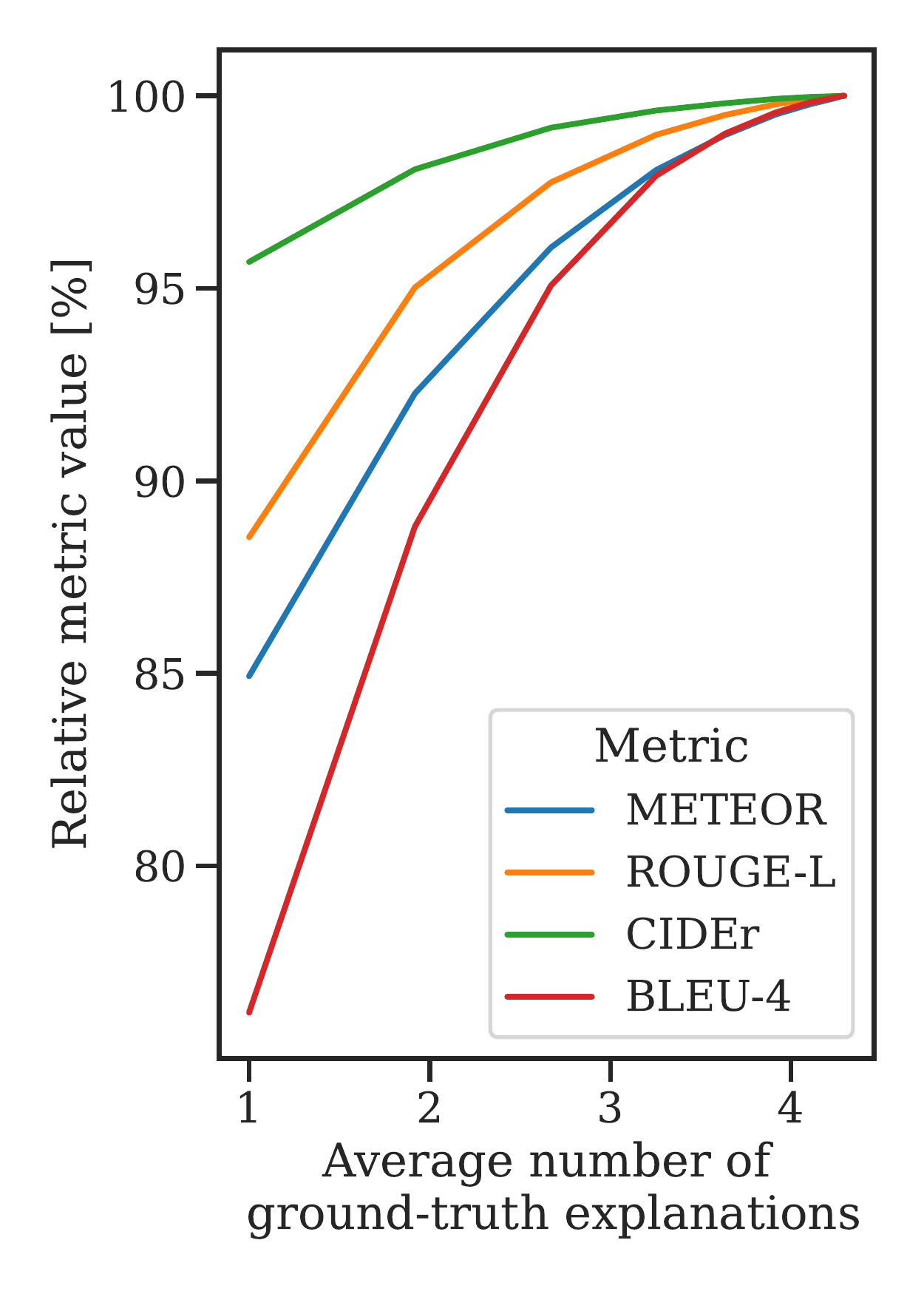}
    \end{subfigure}
    \hfill
    \caption{%
    Explanation generation results for PJ-X on the CLEVR-X test set according to question (left) and answer (middle) types \rev{compared to the overall explanation quality}. Easier types yield higher METEOR scores.
    NLG metrics using different numbers of ground-truth explanations on the CLEVR-X test set (right). CIDEr converges faster than the other NLG metrics.}%
    \label{fig:clevrx-answer-question-type-full-results-and-metrics-with-different-gt}%
\end{figure}

In Fig.~\ref{fig:clevrx-answer-question-type-full-results-and-metrics-with-different-gt} (left and middle), we present the performance for PJ-X on CLEVR-X for the 9 question and 3 answer types. The explanation results for samples which require counting abilities (counting answers) are lower than those for attribute answers (57.3 vs.\ 63.3).
This is in line with prior findings that VQA models struggle with counting problems~\cite{Trott2018InterpretableCF}.
\rev{The explanation quality for binary questions is even lower with a METEOR score of only 55.6.}
The generated explanations are of higher quality for easier question types; \textit{zero-hop} questions yield a METEOR score of 64.9 compared to 62.1 for \textit{three-hop} questions. 
It can also be seen that \textit{single-or} questions are harder to explain than \textit{single-and} questions.
These trends can be observed across all NL\rev{G} explanation metrics.

\subsection{Influence of using different numbers of ground-truth explanations}\label{sec:length-results}
In this section, we 
study the influence of using multiple ground-truth explanations for evaluation on the behavior of the NL\rev{G} metrics. 
This gives insights about whether the metrics can correctly rate a model's performance with a limited number of ground-truth explanations.
We set an upper bound $k$ on the number of explanations used and randomly sample $k$ explanations if a test sample has more than $k$ explanations for $k \in\{1,2,\dots,10\}$.
Figure~\ref{fig:clevrx-answer-question-type-full-results-and-metrics-with-different-gt} (right) shows the NL\rev{G} metrics (normalized with the maximum value for each metric on the test set for all ground-truth explanations) for the PJ-X model depending on the average number of ground-truth references used on the test set.

Out of the four metrics, BLEU-4 converges the slowest, requiring close to 3 ground-truth explanations to obtain a relative metric value of 95\%. Hence, BLEU-4 might not be able to reliably predict the explanation quality on the e-SNLI-VE dataset which has only one explanation for each test sample.
CIDEr converges faster than ROUGE and METEOR, and achieves 95.7\% of its final value with only one ground-truth explanation.
\rev{This could be caused by the fact, that CIDEr utilizes a tf-idf weighting scheme for different words, which is built from all reference sentences in the subset that the metric is computed on. This allows CIDEr to be more sensitive to important words (e.g.\ attributes and shapes) and to give less weight, for instance, to stopwords, such as \enquote{the}.}
The VQA-X and e-SNLI-VE datasets contain much lower average numbers of explanations for each dataset sample (1.4 and 1.0). Since there could be many more possible explanations for samples in those datasets that describe different aspects than those mentioned in the ground truth, automated metric may not be able to correctly judge a prediction \rev{even if it is correct and faithful w.r.t.\ to the image and question.
}

\begin{figure}[t]
    \centering
    \begin{subfigure}[t]{0.31\textwidth}
        \centering
        \scriptsize
        \caption*{\textbf{Question}: How many tiny red things are the same material as the big sphere?}
        \includegraphics[width=.75\linewidth]{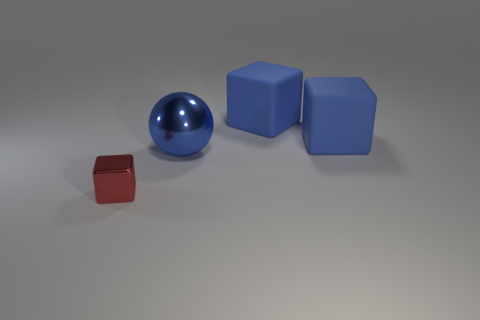}
        \caption*{%
        \textbf{GT Answer $\mid$ Explanation}:\\
        \rev{1} $\mid$ The tiny red metal block has the same material as a big sphere.\\
        \textbf{Pred. Answer $\mid$ Expl.}\\
        \rev{1} $\mid$ \rev{There is the tiny red metal block which has the identical material as a big sphere.}\\
        \rev{\textbf{B4 / M / RL / C:}\\100.0 / 100.0 / 100.0 / 744.0}
        
        }%
       \label{fig:clevrx-qualitative-example-1}
    \end{subfigure}
    \hfill
    \begin{subfigure}[t]{0.31\textwidth}
        \centering
        \caption*{\textbf{Question}: The cylinder has what size?\\}
        \includegraphics[width=.75\linewidth]{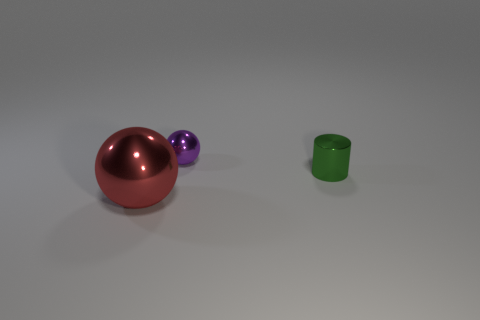}
        \caption*{%
        \textbf{GT Answer $\mid$ Explanation}:\\
        Small $\mid$ The cylinder is small.\\\\\\
        \textbf{Pred. Answer $\mid$ Expl.}\\
        Small $\mid$ \rev{The cylinder is tiny}.\\\\\\
        \rev{\textbf{B4 / M / RL / C:}\\100.0 / 100.0 / 100.0 / 462.4}
        }%
        \label{ffig:clevrx-qualitative-example-2}
    \end{subfigure}
    \hfill
    \begin{subfigure}[t]{0.31\textwidth}
        \centering
        \caption*{\textbf{Question}: Are there any small matte cubes?\\}
        \includegraphics[width=.75\linewidth]{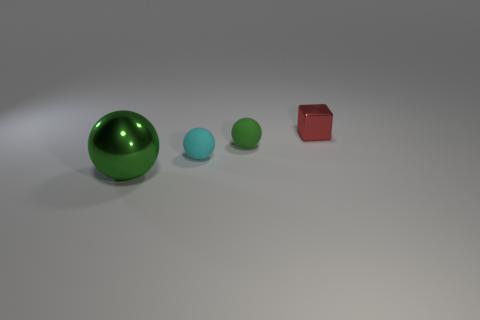}
        \caption*{%
        \textbf{GT Answer $\mid$ Explanation}:\\
        No $\mid$ There are no small matte cubes.\\\\
        \textbf{Pred. Answer $\mid$ Expl.}\\
        Yes $\mid$ There is a small matte cube.\\\\
        \rev{\textbf{B4 / M / RL / C:}\\0.0 / 76.9 / 57.1 / 157.1}
        }%
        \label{ffig:clevrx-qualitative-example-3}
    \end{subfigure}
    \caption{Examples for answers and explanations generated with the PJ-X framework on the CLEVR-X dataset, showing correct answer predictions (left, middle) and a failure case (right). \rev{The NLG metrics obtained with the explanations for the correctly predicted answers are high compared to those for the explanation corresponding to the wrong answer prediction.}
    }%
    \label{fig:clevrx-qualitative-examples}%
\end{figure}

\subsection{Qualitative explanation generation results}\label{sec:quali-results}
We show examples for explanations generated with the PJ-X framework on CLEVR-X in Fig.~\ref{fig:clevrx-qualitative-examples}. As can be seen across the three examples presented, PJ-X generates high-quality explanations which closely match the ground-truth explanations.

In the left-most example in Fig.~\ref{fig:clevrx-qualitative-examples}, we can observe slight variations in grammar when comparing the generated explanation to the ground-truth explanation. However, the content of the generated explanation corresponds to the ground truth. 
Furthermore, some predicted explanations differ from the ground-truth explanation in the use of another synonym for a predicted attribute. For instance, in the middle example in Fig.~\ref{fig:clevrx-qualitative-examples}, the ground-truth explanation describes the size of the cylinder as ``small'', whereas the predicted explanation uses the equivalent attribute ``tiny''.
In contrast to other datasets, the set of ground-truth explanations for each sample in CLEVR-X contains these variations. Therefore, the automated NLG metrics \rev{do} not decrease when such variations are found in the predictions. \rev{For the first and second example, PJ-X obtains the highest possible explanation score (100.0) in terms of the BLEU-4, METEOR, and ROUGE-L metrics.}

We show a failure case where PJ-X predicted the wrong answer in Fig.~\ref{fig:clevrx-qualitative-examples}~(right). The generated answer-explanation pair shows that the predicted explanation is consistent with the wrong answer prediction and does not match the input question-image pair.
\rev{The NLG metrics for this case are significantly weaker with a BLEU-4 score of $0.0$, as there are no matching $4$-grams between the prediction and the ground truth.}

\section{Conclusion}
We introduced the novel \clevrx{} dataset which contains natural language explanations for the VQA task on the CLEVR dataset.
Our user study confirms that the explanations in the \clevrx{} dataset are complete and match the questions and images.
Furthermore, we have provided baseline performances using the PJ-X and FM frameworks on the \clevrx{} dataset. The structured nature of our proposed dataset allowed the detailed evaluation of the explanation generation quality according to answer and question types. We observed that the generated explanations were of higher quality for easier answer and question categories.
One of our findings is, that explanations for counting problems are worse than for other answer types, suggesting that further research into this direction is needed.
Additionally, we find that the four NL\rev{G} metrics used to evaluate the quality of the generated explanations exhibit different convergence patterns depending on the number of available ground-truth references. 

Since this work only considered two natural language generation methods for VQA as baselines, the natural next step will be the benchmarking and closer investigation of additional recent frameworks for textual explanations in the context of VQA on the \clevrx{} dataset.
We hope that our proposed \clevrx{} benchmark will facilitate further research to improve the generation of natural language explanations in the context of vision-language tasks.
\section{Acknowledgements}
The authors thank the Amazon Mechanical Turk workers that participated in the user study. This work was supported by the DFG – EXC number 2064/1 – project number 390727645, by the DFG: SFB 1233, Robust Vision: Inference Principles and Neural Mechanisms - project number: 276693517, by the ERC (853489 - DEXIM), and by the BMBF (FKZ: 01IS18039A). The authors thank the International Max Planck Research School for Intelligent Systems (IMPRS-IS) for supporting Leonard Salewski.

%
\bibliographystyle{splncs04}
\bibliography{mybib.bib}
\end{document}